\documentclass{article}

\usepackage[nonatbib, preprint]{neurips_2025}

\usepackage{biblatex}
\usepackage[utf8]{inputenc} 
\usepackage[T1]{fontenc}    
\usepackage{hyperref}       
\usepackage{url}            
\usepackage{booktabs}       
\usepackage{amsfonts}       
\usepackage{nicefrac}       
\usepackage{microtype}      
\usepackage{xcolor}         
\usepackage{amsmath}

\usepackage{multirow}
\usepackage{graphicx}
\usepackage{pifont}

\title{Pro-AD: Learning Comprehensive Prototypes with Prototype-based Constraint for Multi-class Unsupervised Anomaly Detection}

\author{
    Ziqing Zhou$^{\dagger,1}$ \\ 
    $^1$Fudan University\\ 
    \texttt{zqzhou23@m.fudan.edu.cn} \\ 
    \And 
    Yurui Pan$^1$ \\ 
    $^1$Fudan University \\ 
    \texttt{yrpan24@m.fudan.edu.cn} \\ 
    \And 
    Lidong Wang$^1$ \\ 
    $^1$Fudan University \\ 
    \texttt{ldwang22@m.fudan.edu.cn} \\ 
    \And 
    Wenbing Zhu$^{1, 5}$ \\ 
    $^1$Fudan University \\ 
    $^5$Rongcheer Co., Ltd. \\ 
    \texttt{wbzhu23@m.fudan.edu.cn} \\ 
    \And 
    Mingmin Chi$^{1,*}$ \\ 
    $^1$Fudan University \\ 
    \texttt{mmchi@fudan.edu.cn} \\ 
    \And 
    Dong Wu$^{6,*}$ \\ 
    $^1$Fudan University \\ 
    \texttt{dwu19@m.fudan.edu.cn} \\ 
    \And 
    Bo Peng$^{3,*}$ \\ 
    $^3$ Shanghai Ocean University \\ 
    \texttt{bpeng@shou.edu.cn} \\ 
}

\begin{document}

\maketitle

\begin{abstract}
Prototype-based reconstruction methods for unsupervised anomaly detection utilize a limited set of learnable prototypes which only aggregates insufficient normal information, resulting in undesirable reconstruction. However, increasing the number of prototypes may lead to anomalies being well reconstructed through the attention mechanism, which we refer to as the "Soft Identity Mapping" problem. In this paper, we propose Pro-AD to address these issues and fully utilize the prototypes to boost the performance of anomaly detection. Specifically, we first introduce an expanded set of learnable prototypes to provide sufficient capacity for semantic information. Then we employ a Dynamic Bidirectional Decoder which integrates the process of the normal information aggregation and the target feature reconstruction via prototypes, with the aim of allowing the prototypes to aggregate more comprehensive normal semantic information from different levels of the image features and the target feature reconstruction to not only utilize its contextual information but also dynamically leverage the learned comprehensive prototypes. Additionally, to prevent the anomalies from being well reconstructed using sufficient semantic information through the attention mechanism, Pro-AD introduces a Prototype-based Constraint that applied within the target feature reconstruction process of the decoder, which further improves the performance of our approach. Extensive experiments on multiple challenging benchmarks demonstrate that our Pro-AD achieve state-of-the-art performance, highlighting its superior robustness and practical effectiveness for Multi-class Unsupervised Anomaly Detection task.
\end{abstract}

\section{Introduction}

Multi-class Unsupervised Anomaly Detection (MUAD) \cite{uniad} represents a critical and challenging frontier in computer vision, with profound implications for real-world applications such as industrial quality control. The core objective is to develop a single, versatile model capable of identifying deviations from normality across a diverse range of object or texture classes, relying solely on anomaly-free training data. Previous methods \cite{uniad, omnial, hvq-trans, mambaad, dinomaly} trained their models using the anomalous-free features that extracted from training set images by the pre-trained backbone, with the expectation that the models would learn the distribution of normality from normal features, thereby enabling the identification of out-of-distribution anomalies. However, as the number of class increases, it becomes increasingly difficult to learn all the distribution of normal samples for all classes, leading to a decline in the model's detection performance.

\begin{figure}
    \centering
    \includegraphics[width=1.0\linewidth]{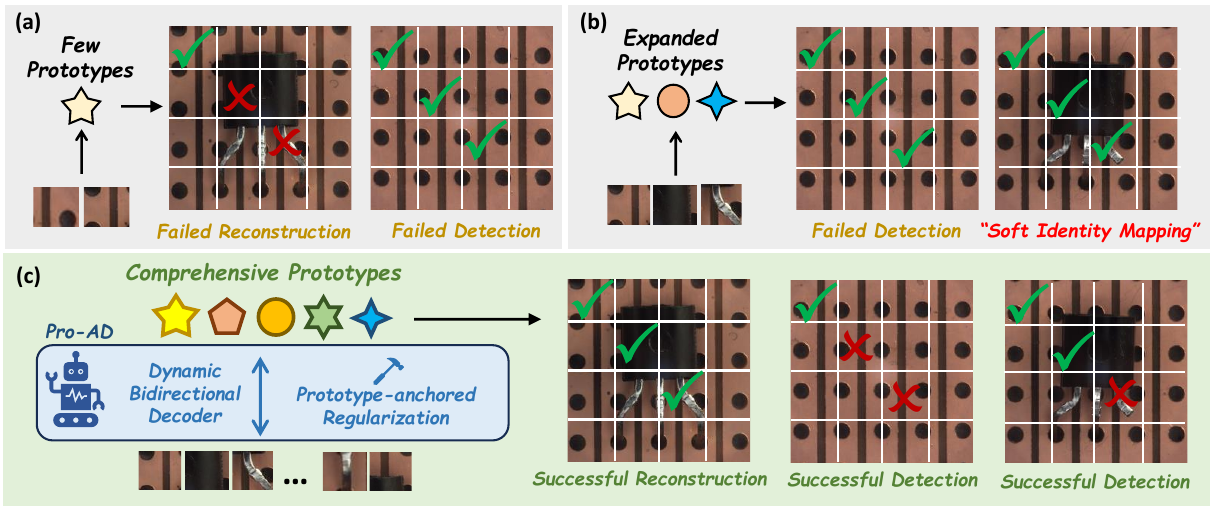}
    \caption{\textbf{Illustration of the "Soft Identity Mapping" problem}. (a) Few prototypes can only aggregation background related features with failed reconstruction and detection. (b) Expanded prototypes fail to detection certain anomalies as well and may cause "Soft Identity Mapping" problem. (c) Our Pro-AD learns a set of comprehensive prototypes and achieves both successful reconstruction and detection.}
    \label{fig:moti}
\end{figure}

Some methods have been attempted to use prototype learning \cite{prototypelearning, prototype2} strategy to reduce the difficulty of covering normal distribution for multi-class anomaly detection. Patchcore \cite{patchcore}, for example, utilizes pre-trained Convolutional Neural Network (CNN) \cite{wideresnet} backbone to extract patch features from the training set images as normal prototypes, which are then aggregated into a "memory bank" of normal prototypes. During inference, anomalies are identified by measuring the feature distance between a test patch and its nearest neighbors in this memory bank. While intuitively appealing for simplifying the learning task, these approaches face practical limitations. The memory footprint can become prohibitively large as the number of classes and the volume of training data increase. Furthermore, a potential misalignment can arise between these pre-stored patch features and the features of test images, which may not always be optimally discriminative, leading to suboptimal anomaly detection performance.

More recent methods, such as INP-Former \cite{inp}, attempt to learn a compact set of generic prototypes from all training data using a transformer architecture. The underlying expectation is that these generic prototypes, through the attention mechanism, can reconstruct normal features but not abnormal ones. However, this approach presents a significant dilemma: a small set of prototype are not able to captures abundant normal representations, which may cause failed reconstruction of complex normal feature or failed detection for special structural anomalies as illustrated in Fig. \ref{fig:moti}(a). Conversely, Simply increasing the number of prototypes to aggregate more normal information can help in the complex normal reconstruction, but it still cannot solve the failure in detecting certain logic anomalies and it may lead to the reconstruction of abnormal features, thus losing the ability to identify anomalies. We define this phenomenon as the "Soft Identity Mapping" problem within the Transformer mechanism.


To address this issue, we introduce Pro-AD, a reconstruction-based model consists of a pre-trained encoder for robust feature extraction, an adaptive noisy bottleneck to adaptively add noise into the feature as well as to maintain the consistency of the feature space between the prototype and the target feature, an expanded set of learnable prototypes to aggregate comprehensive normal semantic information for later target reconstruction, and a dynamic bidirectional Decoder with an additional Prototype-based Constraint for both comprehensive prototype learning and discriminative target feature reconstruction. Under the interaction of these modules, our approach can effectively detect both common and difficult anomalies under the guidance of learned normal prototypes. Extensive experiments conducted on the three datasets MVTec-AD \cite{mvtec}, VisA \cite{visa}, and Real-IAD \cite{realiad} have demonstrated that our Pro-AD achieves state-of-the-art performance. Our main contributions are summarized as follows:
\begin{itemize}
\item[$\bullet$] We define the "Soft Identity Mapping" problem for prototype-based anomaly detection methods when increasing the number of prototypes.
\item[$\bullet$] We propose Pro-AD, a reconstruction-based framework that utilizes a dynamic bidirectional decoder with Prototype-based Constraint for both comprehensive prototype learning and robust prototype-guided reconstruction of target feature.
\item[$\bullet$] Our method achieves state-of-the-art performance on three challenging benchmarks, including MVTec \cite{mvtec}, VisA\cite{visa}, and Real-IAD\cite{realiad}, demonstrating its superior robustness and effectiveness in Multi-class Unsupervised Anomaly Detection task.
\end{itemize}

\section{Related Work}
\subsection{Embedding-based Anomaly Detection}
Embedding-based AD methods leverage pre-trained feature extractors to distinguish normal from anomalous patterns. These approaches essentially create reference representations that serve as prototypes of normality, making them conceptually related to our work. \cite{padim, mdnd, dfm} model the distribution of normal features and identifies anomalies through statistical distance measures, while PatchCore \cite{patchcore} constructs a memory bank of normal patch features for nearest-neighbor comparison during inference. Simplenet \cite{simplenet} generates anomaly features by adding Gaussian noise to normal features and then learns a binary discriminator to distinguish anomaly features from normal ones. Despite their effectiveness, these methods are limited by designing structural differences between teacher and student.
 
\subsection{Reconstruction-based Anomaly Detection}
Reconstruction-based methods operate on the principle that normal patterns can be accurately reconstructed while anomalies cannot. This concept has evolved from class-specific models toward unified multi-class frameworks. UniAD \cite{uniad} pioneered a unified reconstruction approach for multiple categories, DeSTSeg \cite{destseg} proposes a denoising knowledge distillation and employs a segmentation network for accurate anomaly localization with synthetic samples. More recent approaches like MambaAD \cite{mambaad} and Dinomaly \cite{dinomaly} leverage advanced architectures such as State Space Models \cite{mamba} and DINO \cite{dinov2} to enhance reconstruction capabilities. reconstruct features extracted from a pre-trained model and achieve state-of-the-art performance for unified anomaly detection. However, pixel-level anomaly segmentation is still unsatisfactory.

\subsection{Prototype-based Anomaly Detection}
Prototype-based methods represent a promising direction that bridges embedding and reconstruction approaches by using reference patterns to guide the anomaly detection process. PatchCore \cite{patchcore} rely on pre-stored normal prototypes extracted from the training set, which can suffer from the misaligned normality problem. Other approaches \cite{memorizing, learning, park2020learning, pixel} incorporate prototypes into the reconstruction process to avoid the identical shortcut issue, but still suffer from the misaligned problem. INP-Former \cite{inp} solves the misalignment problem by learning a small number of prototype tokens for target feature reconstruction. However, it remains poor response to certain logical anomalies.

\begin{figure}
    \centering
    \includegraphics[width=0.99\linewidth]{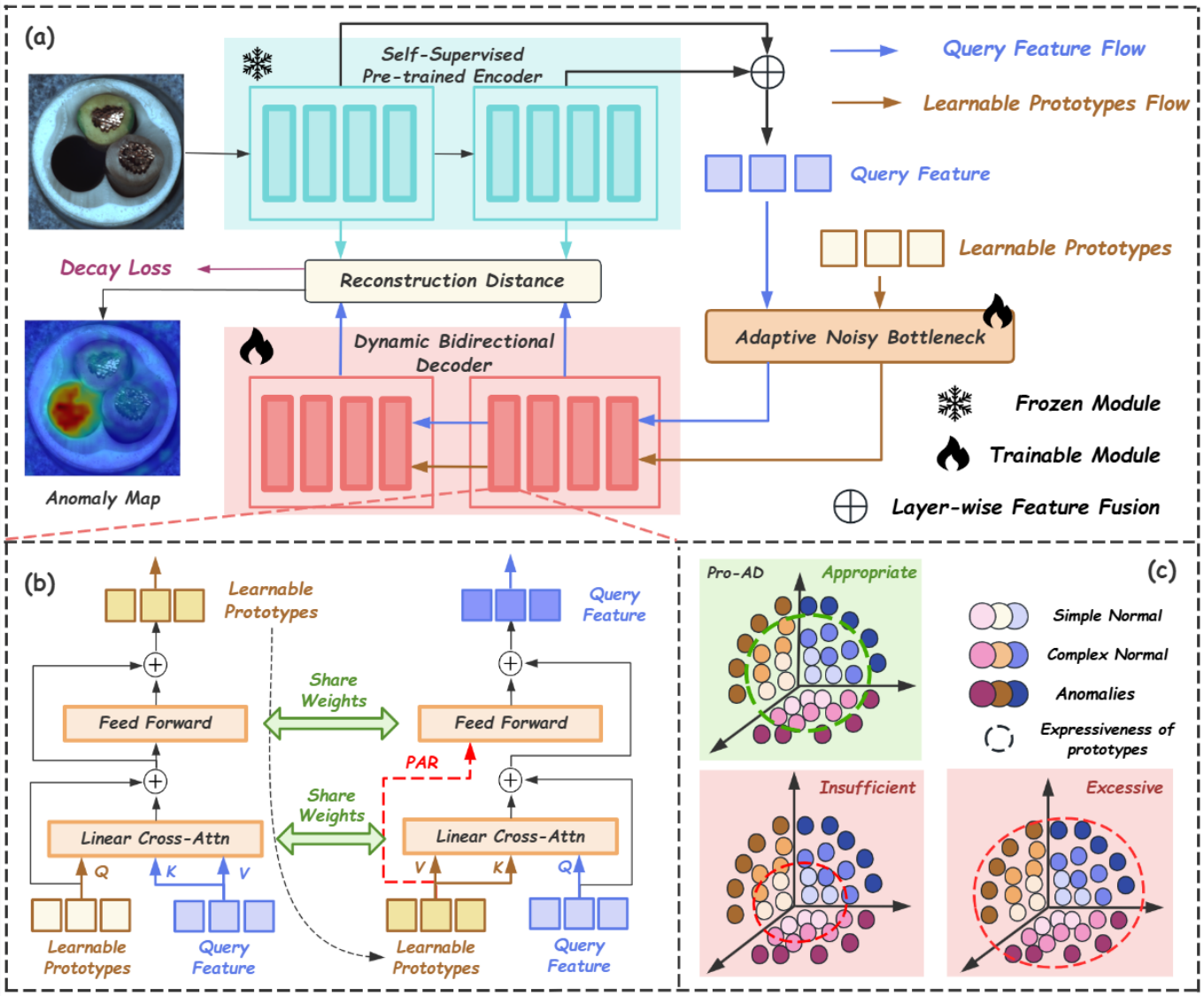}
    \caption{\textbf{Overview of our Pro-AD framwork for anomaly detection.} (a) Our model consists of a pre-trained Encoder, an adaptive noisy bottleneck and a bidirectional Prototype-Guided decoder. (b) Detailed architecture of the bidirectional Prototype-Guided decoder layer. (c) The prototype's ability to express normal and abnormal features.}
    \label{fig:pipeline}
\end{figure}

\section{Methods}
\subsection{Overview}
To learn comprehensive prototypes and to fully exploit the latent of prototypes for robust anomaly detection, we propose Pro-AD, as depicted in Fig. \ref{fig:pipeline}(a). The model is able to learn a set of comprehensive prototypes and to dynamically utilize them for the target feature reconstruction with the application of an Adaptive Noisy Bottleneck and  a Dynamic Bidirectional Decoder with extra Prototype-based Constraint. The linear cross-attention and feed forward network for prototype learning and those for target feature reconstruction are shared-weighted as shown in Fig. \ref{fig:pipeline}(b), which ensures that our method is basically consistent with the parameter count of previous methods, as shown in the Tab.\ref{tab:param}

Given an input image \( I \in \mathbb{R}^{H \times W \times 3} \), we employ the pre-trained backbone as encoder \( E \) to extract multi-scale features \( f_{en} = \{ f_{en}^{1}, ..., f_{en}^{L} \,|\, f_{en}^{L} \in \mathbb{R}^{N \times C},\, N = \frac{HW}{p^{2}} \} \), where \( p \) represents the patch number of the encoder. 
To obtain a comprehensive feature representation, we perform a summation over the features from the second to the ninth encoder layers, resulting in the input features \( F_{I} \in \mathbb{R}^{N \times C} \). This summation enables the integration of both low-level textural features and high-level semantic information, facilitating a richer and more discriminative representation of the input.

Next, we initialize a set of learnable prototype tokens \( P \in \mathbb{R}^{N \times C} \), which shares the same dimensionality as the input features. Unlike previous method \cite{inp} of limiting the number of prototypes as 6 "INPs", our approach enlarge the number of prototypes to the same as patches number, i.e. $N$ prototypes, which allows the prototypes to aggregate comprehensive normal semantic information and effectively capture contextual details in normal feature. Notably, since our prototype is common to all features, it doesn't add much of parameters compared to previous methods \cite{inp, patchcore}.

\subsection{Adaptive Noisy Bottleneck}
Follow Dinomaly \cite{dinomaly}, we employ dropout strategy in bottleneck to add noise to the input features. This process is as follows:
\begin{equation}
\begin{aligned}
    & \hat{Q} = Dropout_{prob} ( GELU (W_1 \cdot F_I + b_1) ), \\
    & Q_{bn} = Dropout_{prob} ( W_2 \cdot \hat{Q} + b_2 ),
\end{aligned}
\end{equation}

where $W_1$ and $b_1$ represent the weight matrix and bias vector of the linear projection layer respectively, and $prob$ denotes the probability of dropout. 
Additionally, since the learnable prototypes require performing aggregation and target reconstruction calculations with the query feature in the decoder, we let the learnable prototypes go through the bottleneck as well:
\begin{equation}
\begin{aligned}
    & \hat{P} = Dropout_{0.0} (GELU (W_1 \cdot P + b_1)), \\
    & P_{bn} = Dropout_{0.0} (W_2 \cdot \hat{P} + b_2) ,
\end{aligned}
\end{equation}
here we set the dropout rate to 0 so that we can add no noise to the prototypes and keep the stability and representativeness of the prototypes. By doing this, we ensure that the learnable prototypes and query features remain in a unified feature space before being fed into the decoder

\begin{table}
    \centering
    \caption{Comparison of parameter scale between INP-Former and our Pro-AD.}
    \begin{tabular}{c|c|c|c|c|c}
    \toprule
     \multirow{2}{*}{Methods $\downarrow$} & \multicolumn{4}{c|}{Essential Components} & \multirow{2}{*}{Total}  \\
     \cmidrule(lr){2-5}
       & $Bottleneck$ & $Agg \ Block$ & $Decoder$ & $Prototypes$ &  \\

    \midrule
    INP-Former & 4,722,432 & 7,087,872 & 56,703,072 & 153,600 & 68,666,976 \\
    \midrule
    Ours (Pro-AD) & 4,722,432 & \textbackslash & 56,702,976 & 605,952 & \textcolor{blue}{62,031,360} \\
    \bottomrule
    \end{tabular}
    \label{tab:param}
\end{table}


\subsection{Dynamic Bidirectional Decoder}
Existing method \cite{inp} separates the process of prototype learning and the target feature reconstruction. And the prototypes after aggregation remain unchanged during the target feature reconstruction. This, on one hand, forbids the prototypes to dynamically aggregate different normal semantic information as the reconstruction features changing in the decoder, and limits the guiding effect of the prototype in the reconstruction of target features on the other hand. 
In contrast, we combine two process mentioned above together in each layer of our Dynamic Bidirectional Decoder. Given a set of prototypes $P$ and target feature $F$, we first perform the prototype learning via a formal linear cross-attention(LCA), which is computed as:
\begin{equation}
\begin{aligned}
LCA(Q_P,K_F,V_F) &= \phi(Q_P)(\phi(K_F^{T})V_F)
\label{eq:LCA_1}
\end{aligned}
\end{equation}
where $Q_P$ denotes the query of attention obtained by prototype $P$ through linear mapping, $K_F$ and $V_F$ denote the key and value obtained by target feature $F$ respectively and $\phi$ denotes the activate function. The reason we use linear attention instead of softmax attention is that while softmax attention increases computational cost, the softmax function will bring a sharp attention map, forcing the prototypes to focus only on the important regions of the images and ignore the texture details. On the contrary, by replacing softmax with activate function, linear attention not only reduces the computational cost but also makes the attention map smoother, which directly allows the prototype to globally aggregate detail contextual information from various parts of the target feature. And the specific benefit of this for anomaly detection tasks is that it enables the model to focus on a wider context rather than just a local salient area, thereby improving the ability to detect anomalies in certain contexts. Then, similar to a standard transformer block, the output of the linear cross-attention is processed through a feedforward network (FFN) to obtain the updated prototypes of current decoder layer, as represented in the following process:
\begin{equation}
\begin{aligned}
Attn^{l}_{PQ} &= P^{l} + LCA(P^{l},\ Q^{l},\ Q^{l}),
\end{aligned}
\end{equation}
\begin{equation}
\begin{aligned}
P^{l+1} &= Attn^{l}_{PQ} + FFN(Attn^{l}_{PQ}),
\end{aligned}
\end{equation}
where $l$ denotes the $l$-th layer of the decoder, $Q^l$ denotes the target feature, $P^l$ denotes the learnable prototypes and $P^{l+1}$ denotes the updated prototypes which will be passed to the next layer of decoder. Note that $Q^0 =Q_{bn}$ and $P^0 =P_{bn}$. Next we employ the linear cross-attention for the target feature reconstruction stage as well:
\begin{equation}
\begin{aligned}
LCA(Q_F,K_P,V_P) &= \phi(Q_F)(\phi(K_P^{T})V_P) \\
&=W^TV_F,
\label{eq:LCA_2}
\end{aligned}
\end{equation}
where $Q_F$ denotes the query obtained by target feature $F$, $K_P$ and $V_P$ denote the key and value obtained by the prototypes $P$, $W$ denotes the attention map. The linear attention here is expected to prevent the identity mapping problem \cite{dinomaly}. We rewrite this formula in the following form:
\begin{equation}
\begin{aligned}
LCA(Q_F,K_P,V_P) &=W^TV_F \\
&=Proj(V_F),
\label{eq:rewrite}
\end{aligned}
\end{equation}
as illustrated in Formula (\ref{eq:rewrite}), the calculation of attention can be viewed as a projection of the prototypes. Obviously, the difficulty of reconstructing target features, or in other words, the expressive capability of the prototypes is inversely related to the dimensions of the prototype: a small number of prototypes will lead to unsatisfactory target feature reconstruction while an sufficiently large set of prototypes with strong expressiveness makes it possible for any reconstruction of target feature $F$,resulting in the "Soft Identity Mapping" problem where anomalous feature can also be well reconstructed through the prototypes, leading to a decline of anomaly detection performance.

\subsection{Prototype-based Constraint}
Based on the above analysis, we expect to introduce additional constraints to limit the expressive capability of the prototype within an appropriate range. Same to the prototype learning stage, we first perform a formal linear cross-attention to use the updated prototypes $P^{l+1}$ to reconstruct the target feature at the $l$-th layer of decoder. This process can be written as:
\begin{equation}
\begin{aligned}
f^l_{rec} &= Q^l + LCA(Q^{l},\ P^{l+1},\ P^{l+1}).
\label{eq:naive recon}
\end{aligned}
\end{equation}
We observe that each position of $f^l_{rec}$ is directly related to the weighted combination of all the prototype tokens $P^{l+1}$ in this calculation process. We believe that simply performing a weighted combination is a very loose constraint because it does not restrict the selection of inputs or the range of outputs. And a loose constraint will make it possible to output the compensation information that helps $f^l_{rec}$ to approach $f^l_{en}$, even if there are anomalies in the input $Q^l$, which ultimately leads to the "Soft Mapping Problem". We eventually solved this problem by introducing an additionally tighter constraint, i.e. Prototype-based Constraint, to constrain the output. Specifically, we input the updated prototypes $P^{l+1}$ into a feed forward network to obtain the constrained information $P^{l+1}_{reg}$ at each position. And then superimpose this constrained information on the naive reconstruction results $f^l_{rec}$ to obtain the final target feature reconstruction result $f^l_{D}$:
\begin{equation}
\begin{aligned}
P^{l+1}_{reg} &= FFN(P^{l+1}), \\
f^l_{D} &= f^l_{rec} + P^{l+1}_{reg}.
\label{eq:hard_constraint}
\end{aligned}
\end{equation}
This prototype-based constraint approach, as defined in Equation (\ref{eq:hard_constraint}), fundamentally alters the learning dynamics. It implies that the initial reconstruction component, $f^l_{rec}$, is implicitly tasked with generating features that are complementary to the constrained term $P^{l+1}_{reg}$; In effect, its objective becomes reconstructing $f^l_{en} - P^{l+1}_{reg}$. This is a critical change because it prevents the network from forming anomalies through simple weighted combinations of prototypes. Instead, the reconstruction of each position in the target feature is dually guided: it relies on the contextual information learned through weighted aggregation in $f^l_{rec}$ while also being directly constrained by the explicit normal prototype information $P^l_{reg}$ specific to that position. This dual constraint effectively disrupts the undesired "Soft Identity Mapping," thereby significantly enhancing the model's ability to distinguish anomalies and improving overall performance."

\subsection{Loss}
Inspired by \cite{focal, dinomaly, inp}, we aim to focus more on important regions that are challenging to reconstruct, and less on regions that are simpler to reconstruct, thus we propose a Distance-weighted Decay Loss. This loss mechanism dynamically weighs different spatial locations based on their reconstruction difficulty. Given the encoder features $f^{l}_{E}$ and decoder reconstructions $f^{l}_{D}$ at layer $l$, we first calculate the regional distance map $d^{l}$ using cosine distance, and then we calculate the decay factor $\alpha^l$:
\begin{equation}
\begin{aligned}
& d^{l} = 1 - \frac{(f_E^l)^T \cdot f_D^l}{\Vert f_E^l \Vert \cdot \Vert f_D^l \Vert} , \\
& \alpha^{l} = \frac{d^{l}}{Avg(d^{l})} \\
\end{aligned}
\end{equation}

The overall loss function are computed as:
\begin{equation}
\begin{aligned}
& \mathcal{L} =  \frac{1}{L}\sum_{l=1}^{L} \left[ 1 - \frac{T(f_E^l)^T \cdot T(f_D^l)}{\Vert f_E^l \Vert \cdot \Vert f_D^l \Vert} \right] \\
& \hat{G}^l_{D}(h, w) = gd(G^l_{D}(h, w), \alpha^l(h, w)^ \tau)
\end{aligned}
\end{equation}

where $F(\cdot)$ denotes the flatten operation, $gd(\cdot,\cdot)$ denotes the gradient decay function, ${G}^l_{D}(h, w)$ denotes the gradient of $f_D^l$ at position $(h, w)$ and $\tau$ denotes the hyper parameter. 





\section{Experiments}

\subsection{Experimental Settings}
\textbf{Datasets:} To comprehensively evaluate the effectiveness of our proposed Pro-AD framework, we conduct extensive experiments on three widely-used industrial anomaly detection datasets: \textbf{MVTec-AD} \cite{mvtec}: A benchmark dataset containing 15 categories of industrial objects and textures, comprising 3,629 normal images for training, and 1,982 anomalous images with 498 normal images for testing. It features diverse anomaly types including scratches, dents, contamination, and structural defects. \textbf{VisA} \cite{visa}: A more recent dataset with 12 object categories, containing 8,659 normal images for training and 962 normal images along with 1,200 anomalous images for testing. VisA offers higher resolution and more complex background variations than MVTec-AD. \textbf{Real-IAD} \cite{realiad}: The largest and most challenging industrial anomaly detection dataset to date, featuring 30 different objects captured under various viewpoints. It contains 36,645 normal images for training and a substantial test set with 63,256 normal images and 51,329 anomalous images. This dataset closely resembles real-world industrial inspection scenarios.

\textbf{Evaluation Metrics:} Following standard evaluation protocols in previous anomaly detection works \cite{mambaad, dinomaly, inp}, we use the Area Under the Receiver Operating Characteristic Curve (AUROC), Average Precision (AP), and F1-score-max (F1 max) to evaluate anomaly detection and localization. For anomaly localization specifically, we use Area Under the Per-Region-Overlap (AUPRO) as an additional metric.

\textbf{Implementation Details:} All experiments are implemented in PyTorch. Pro-AD adopts DINO2-R-Base/14 \cite{dinov2} as the default pre-trained encoder. The layer number of the Dynamic Bidirectional Decoder is eight.  All input images are resized to $448 \times 448$ pixels and then center-cropped to $392 \times 392$ pixels to ensure consistent spatial dimensions. The hyperparameters $\tau$ is set to 3.0. The drop rate for MVTec-AD, VisA and Real-IAD are set to $0.2$, $0.3$, $0.4$ respectively. We use Stable AdamW optimizer with an initial learning rate of 1e-4 and weight decay of 1e-5. The model is trained for 200 epochs with a batch size of 16, using a cosine learning rate schedule with 10 warm-up epochs.

\begin{table}
  \caption{Quantitative comparison of anomaly detection and localization performance on the \textbf{MVTec-AD} ~\cite{mvtec} dataset.}
  
  \label{mvtec-table}
  \centering
  \begin{tabular}{c|ccc|cccc}
     \toprule
     \multirow{2}{*}{Methods} & \multicolumn{3}{c|}{Image-Level} & \multicolumn{4}{c}{Pixel-Level} \\
     \cmidrule(lr){2-4} \cmidrule(lr){5-8}
     & \multicolumn{3}{c|}{AUROC / AP / F1\_max} & \multicolumn{4}{c}{AUROC / AP / F1\_max / PRO} \\
     \midrule
     RD4AD \cite{rd4ad}         & \multicolumn{3}{c|}{94.6 / 96.5 / 95.2} & \multicolumn{4}{c}{96.1 / 48.6 / 53.8 / 91.1}   \\
     UniAD \cite{uniad}         & \multicolumn{3}{c|}{96.5 / 98.8 / 96.2} & \multicolumn{4}{c}{96.8 / 43.4 / 49.5 / 90.7}   \\
     SimpleNet \cite{simplenet} & \multicolumn{3}{c|}{95.3 / 98.4 / 95.8} & \multicolumn{4}{c}{96.9 / 45.9 / 49.7 / 86.5}   \\
     DeSTSeg \cite{destseg}     & \multicolumn{3}{c|}{89.2 / 95.5 / 91.6} & \multicolumn{4}{c}{93.1 / 54.3 / 50.9 / 64.8}   \\
     DiAD \cite{diad}           & \multicolumn{3}{c|}{97.2 / 99.0 / 96.5} & \multicolumn{4}{c}{96.8 / 52.6 / 55.5 / 90.7}   \\
     MambaAD \cite{mambaad}     & \multicolumn{3}{c|}{98.6 / 99.6 / 97.8} & \multicolumn{4}{c}{97.7 / 56.3 / 59.2 / 93.1}   \\
     
     Dinomaly \cite{dinomaly}
     & \multicolumn{3}{c|}{99.6 / 99.8 / 99.0} 
     & \multicolumn{4}{c}{98.4 / 69.3 / 69.2 / 94.8}   \\

     INP-Former \cite{inp}
     & \multicolumn{3}{c|}{\underline{99.7} / \underline{99.9} / \underline{99.2}} 
     & \multicolumn{4}{c}{\underline{98.5} / \underline{71.0} / \underline{69.7} / \underline{94.9}}   \\

    \midrule
    Ours (Pro-AD)
     & \multicolumn{3}{c|}{\textbf{99.8} / \textbf{100.0} / \textbf{99.5}}
     & \multicolumn{4}{c}{\textbf{98.8} / \textbf{75.6} / \textbf{72.6} / \textbf{96.4}}   \\
    \bottomrule
  \end{tabular}
\end{table}

\begin{table}
  \caption{Quantitative comparison of anomaly detection and localization performance on the \textbf{VisA} ~\cite{visa} dataset.}
  
  \label{visa-table}
  \centering
  \begin{tabular}{c|ccc|cccc}
     \toprule
     \multirow{2}{*}{Methods} & \multicolumn{3}{c|}{Image-Level} & \multicolumn{4}{c}{Pixel-Level} \\
     \cmidrule(lr){2-4} \cmidrule(lr){5-8}
     & \multicolumn{3}{c|}{AUROC / AP / F1\_max} & \multicolumn{4}{c}{AUROC / AP / F1\_max / PRO} \\
     \midrule
     RD4AD \cite{rd4ad}         & \multicolumn{3}{c|}{92.4 / 92.4 / 89.6} & \multicolumn{4}{c}{98.1 / 38.0 / 42.6 / 91.8}   \\
     UniAD \cite{uniad}         & \multicolumn{3}{c|}{88.8 / 90.8 / 81.8} & \multicolumn{4}{c}{98.3 / 33.7 / 39.0 / 85.5}   \\
     SimpleNet \cite{simplenet} & \multicolumn{3}{c|}{87.2 / 87.0 / 81.8} & \multicolumn{4}{c}{96.8 / 34.7 / 37.8 / 81.4}   \\
     DeSTSeg \cite{destseg}     & \multicolumn{3}{c|}{88.9 / 89.0 / 85.2} & \multicolumn{4}{c}{96.1 / 39.6 / 43.4 / 67.4}   \\
     DiAD \cite{diad}           & \multicolumn{3}{c|}{86.8 / 88.3 / 85.1} & \multicolumn{4}{c}{96.0 / 26.1 / 33.0 / 75.2}   \\
     MambaAD \cite{mambaad}     & \multicolumn{3}{c|}{94.3 / 94.5 / 89.4} & \multicolumn{4}{c}{98.5 / 39.4 / 44.0 / 91.0}   \\
     
     Dinomaly \cite{dinomaly}
     & \multicolumn{3}{c|}{98.7 / 98.9 / 96.2} 
     & \multicolumn{4}{c}{98.7 / \underline{53.2} / \textbf{55.7} / \underline{94.5}}   \\

     INP-Former \cite{inp}
     & \multicolumn{3}{c|}{\underline{98.9} / \underline{99.0} / \underline{96.6} } 
     & \multicolumn{4}{c}{\underline{98.9} / 51.2 / 54.7/ 94.4}   \\

    \midrule
     Ours (Pro-AD)
     & \multicolumn{3}{c|}{\textbf{99.1} / \textbf{99.3} / \textbf{97.0}}
     & \multicolumn{4}{c}{\textbf{99.1} / \textbf{54.1} / \underline{55.4}/ \textbf{95.0}}   \\
    \bottomrule
  \end{tabular}
\end{table}

\begin{table}
  \caption{Quantitative comparison of anomaly detection and localization performance on the \textbf{Real-IAD} ~\cite{realiad} dataset.}
  
  \label{realiad-table}
  \centering
  \begin{tabular}{c|ccc|cccc}
     \toprule
     \multirow{2}{*}{Methods} & \multicolumn{3}{c|}{Image-Level} & \multicolumn{4}{c}{Pixel-Level} \\
     \cmidrule(lr){2-4} \cmidrule(lr){5-8}
     & \multicolumn{3}{c|}{AUROC / AP / F1\_max} & \multicolumn{4}{c}{AUROC / AP / F1\_max / PRO} \\
     \midrule
     RD4AD \cite{rd4ad}         & \multicolumn{3}{c|}{82.4 / 79.0 / 73.9} & \multicolumn{4}{c}{97.3 / 25.0 / 32.7 / 89.6}   \\
     UniAD \cite{uniad}         & \multicolumn{3}{c|}{83.0 / 80.9 / 74.3} & \multicolumn{4}{c}{97.3 / 21.1 / 29.2 / 86.7}   \\
     SimpleNet \cite{simplenet} & \multicolumn{3}{c|}{57.2 / 53.4 / 61.5} & \multicolumn{4}{c}{75.7 / 2.8 / 6.5 / 39.0}   \\
     DeSTSeg \cite{destseg}     & \multicolumn{3}{c|}{82.3 / 79.2 / 73.2} & \multicolumn{4}{c}{94.6 / 37.9 / 41.7 / 40.6}   \\
     DiAD \cite{diad}           & \multicolumn{3}{c|}{75.6 / 66.4 / 69.9} & \multicolumn{4}{c}{88.0 / 2.9 / 7.1 / 58.1}   \\
     MambaAD \cite{mambaad}     & \multicolumn{3}{c|}{86.3 / 84.6 / 77.0} & \multicolumn{4}{c}{98.5 / 33.0 / 38.7 / 90.5}   \\
     
     Dinomaly \cite{dinomaly}
     & \multicolumn{3}{c|}{89.3 / 86.8 / 80.2} 
     & \multicolumn{4}{c}{98.8 / 42.8 / 47.1 / 93.9}   \\

     INP-Former \cite{inp}
     & \multicolumn{3}{c|}{\underline{90.5} / \underline{88.1} / \underline{81.5}} 
     & \multicolumn{4}{c}{\underline{99.0} / \textbf{47.5} / \textbf{50.3} / \underline{95.0}}   \\

    \midrule
    Ours (Pro-AD)
     & \multicolumn{3}{c|}{\textbf{91.5} / \textbf{88.9} / \textbf{83.0}}
     & \multicolumn{4}{c}{\textbf{99.2} / \textbf{47.5} / \underline{49.6} / \textbf{95.3}}   \\
    \bottomrule
  \end{tabular}
\end{table}

  
     
     

  
     

\subsection{Main Results}
We compare the proposed Pro-AD with several stateof-the-art (SOTA) methods for multi-class anomaly detection, including reconstruction-based methods RD4AD \cite{rd4ad}, UniAD \cite{uniad}, DiAD \cite{diad}, MambaAD \cite{mambaad}, Dinomaly \cite{dinomaly} and INP-Former \cite{inp}, and embedding-based methods SimpleNet \cite{simplenet} and DeSTSeg\cite{destseg}.

The experimental results on the MVTec-AD datasets are presented in Tab \ref{mvtec-table}. Our method achieves SOTA performance, with image-level metrics of \textbf{99.8}/\textbf{100.0}/\textbf{99.5} and pixel-level metrics of \textbf{98.8}/\textbf{75.6}/\textbf{72.6}/\textbf{96.4}.
Compared to the second-best results, our method improves by \textcolor{red}{0.1$\uparrow$} / \textcolor{red}{0.2$\uparrow$} / \textcolor{red}{0.4$\uparrow$} at the image level and
by \textcolor{red}{0.2$\uparrow$} / \textcolor{red}{4.6$\uparrow$} / \textcolor{red}{2.9$\uparrow$} / \textcolor{red}{1.5$\uparrow$} at the pixel level.

The experimental results on the VisA datasets are presented in Tab \ref{visa-table}. Our method achieved state-of-the-art performance on seven scores and reached the second highest level on the P-F1\_max score, with image-level metrics of \textbf{99.1}/\textbf{99.3}/\textbf{97.0} and pixel-level metrics of \textbf{99.1}/\textbf{54.1}/\underline{55.4}/\textbf{95.0}.
Compared to the second-best results, our method improves by \textcolor{red}{0.2$\uparrow$} / \textcolor{red}{0.3$\uparrow$} / \textcolor{red}{0.4$\uparrow$} at the image level and
by \textcolor{red}{0.2$\uparrow$} / \textcolor{red}{0.9$\uparrow$} / 0.3$\downarrow$ / \textcolor{red}{0.5$\uparrow$} at the pixel level.

The experimental results on the Real-IAD datasets are presented in Tab \ref{realiad-table}. On the most challenging dataset, our method achieved state-of-the-art performance on most of scores as well, with image-level metrics of \textbf{91.5}/\textbf{88.9}/\textbf{83.0} and pixel-level metrics of \textbf{99.2}/\textbf{47.5}/\underline{49.6}/\textbf{95.3}.
Compared to the second-best results, our method improves by \textcolor{red}{1.0$\uparrow$} / \textcolor{red}{0.8$\uparrow$} / \textcolor{red}{1.5$\uparrow$} at the image level and
by \textcolor{red}{0.2$\uparrow$} / \textcolor{red}{$\sim$} / 0.3$\downarrow$ / \textcolor{red}{0.5$\uparrow$} at the pixel level.
The SOTA performance achieved across the three datasets showcases the effectiveness and robustness of our method. 

\section{Ablation Study}
We conduct experiments on MVTec-AD \cite{mvtec} to validate the effectiveness of the proposed components, i.e., Adaptive Noisy Bottleneck ("ANB"), Dynamic bidirectional Decoder and Prototype-based Constraint, as illustrated in Tab. \ref{tab:ablation}.
With the introduction of Adaptive Noisy Bottleneck, our approach improves by 0.4$\uparrow$ / 0.3$\uparrow$ / 0.3$\uparrow$ at the image level and
by 0.8$\uparrow$ / 3.7$\uparrow$ / 4.8$\uparrow$ / 1.1$\uparrow$ at the pixel level compared with the baseline results. The significant improvement at the pixel level indicates that the "ANB" effectively prevents identity mapping and enhances the model's ability to detect anomalous details.
By transforming the static unidirectional decoder into a dynamic bidirectional decoder, the pixel-level metrics of our method have been further improved by 0.1$\uparrow$ / 4.2$\uparrow$ / 1.0$\uparrow$ / 0.5$\uparrow$, which proves that the introduction of bidirectional information interaction is helpful to the detailed reconstruction as well. Finally, with the application of the Prototype-based Constraint, our Pro-AD gains a greatly improvement on pixel-level anomaly detection again and achieves the state-of-the-art performances on the MVTec-AD dataset, which indicates that the constraints on the reconstruction process effectively reduce the missed detections of anomalies and improve the performance of anomaly detection.

\begin{figure}
    \centering
    \includegraphics[width=0.99\linewidth]{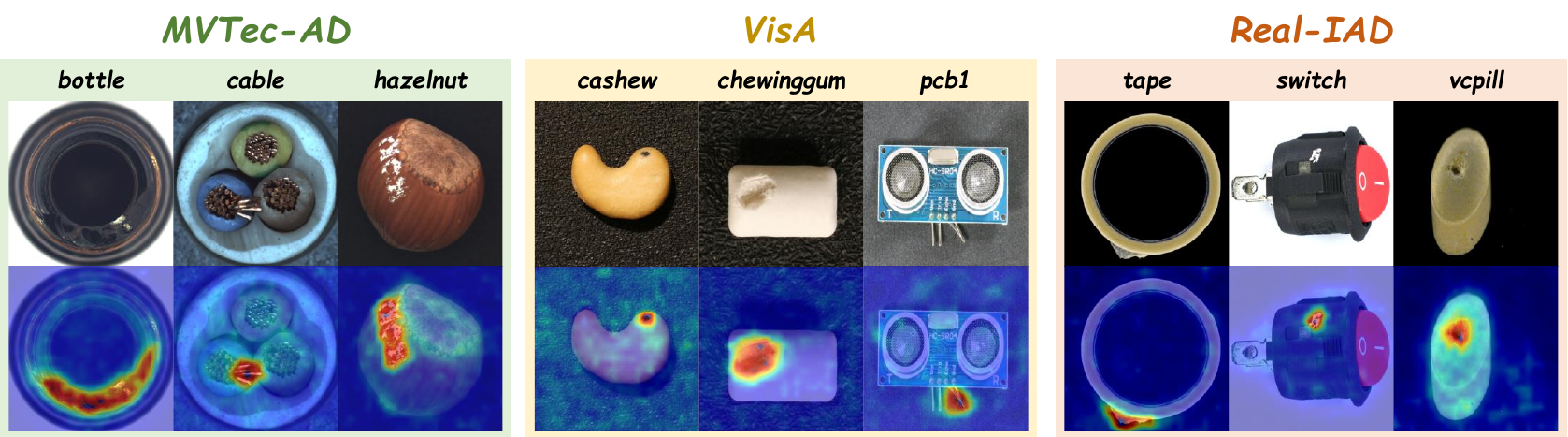}
    \caption{\textbf{Visualization results} of anomaly localization on the MVTec-AD \cite{mvtec}, VisA \cite{visa}, and Real-IAD \cite{realiad} datasets for multi-class anomaly detection. The first row presents the input images, while the second row displays the corresponding anomaly maps}
    \label{fig:vis}
\end{figure}

\begin{table}
    \centering
    \caption{\textbf{Overall ablation} of the essential components of our Pro-AD on MVTec-AD \cite{mvtec} dataset. "ANB" refers to the Adaptive Noisy Bottleneck, "Dynamic" refers to the use of dynamic bidirectional decoder and "PR" stands for Prototype-based Constraint.}
    \begin{tabular}{ccc|ccc|cccc}
     \toprule
     \multirow{2}{*}{"ANB"} & \multirow{2}{*}{"Dynamic"} & \multirow{2}{*}{"PR"} & \multicolumn{3}{c|}{Image-Level} & \multicolumn{4}{c}{Pixel-Level} \\
     \cmidrule(lr){4-6} \cmidrule(lr){7-10}
       & & & \multicolumn{3}{c|}{I-AUROC/I-AP/I-F1\_max} & \multicolumn{4}{c}{P-AUROC/P-AP/P-F1\_max/P-PRO} \\

     \midrule
     \ding{55} & \ding{55} & \ding{55} & \multicolumn{3}{c|}{99.1 / 99.5 / 98.2} & \multicolumn{4}{c}{97.5 / 63.7 / 64.5 / 93.6} \\
     \checkmark & \ding{55} & \ding{55} & \multicolumn{3}{c|}{99.5 / 99.8 / 99.1} & \multicolumn{4}{c}{98.3 / 67.3 / 69.3 / 94.7} \\
     
     \checkmark & \checkmark & \ding{55} & \multicolumn{3}{c|}{99.6 / 99.8 / 99.3} & \multicolumn{4}{c}{98.4 / 71.5 / 70.3 / 95.2} \\
     \checkmark & \checkmark & \checkmark &\multicolumn{3}{c|}{\textbf{99.8 / 100.0 / 99.5}} & \multicolumn{4}{c}{\textbf{98.8 / 75.6 / 72.6 / 96.4}} \\
     \bottomrule
    \end{tabular}
    \label{tab:ablation}
\end{table}

\section{Conclusion}
We propose Pro-AD, a prototype-based reconstruction method for anomaly detection that fully exploit the latent of comprehensive prototypes. By dynamically aggregating normal semantic information into learnable prototypes while simultaneously using these prototypes to guide the reconstruction of target features, Pro-AD significantly enhances anomaly detection performance. The introduction of the Prototype-based Constraint further improve the performance of our approach by resonably limiting the representative abilities of prototypes. Extensive experiments on MVTec-AD, VisA, and Real-IAD datasets demonstrate that Pro-AD achieves SOTA or comparable performance on multi-class anomaly detection tasks. 

\textbf{Limitations and Future Works:} 
Currently our method only support unsupervised setting. And, although our method is able to detect logical anomalies that closely resemble the background distribution, such as the misplaced anomalies in the Transistor class of the MVTecAD dataset, we encounters certain limitations when dealing more challenging logical anomalies within datasets specifically designed for logical anomaly detection tasks, e.g. MVTec-LOCO \cite{loco}. We recon that the Prototype-based Constraint is effective for detecting logical anomalies that are not sensitive to location changes and have a larger area. In future work, we plan to expand Pro-AD to different settings includes few-shot anomaly detection and zero-shot anomaly detection, and further improve our detection performance on logical anomalies.

\printbibliography

@inproceedings{mvtec,
  title={MVTec AD--A comprehensive real-world dataset for unsupervised anomaly detection},
  author={Bergmann, Paul and Fauser, Michael and Sattlegger, David and Steger, Carsten},
  booktitle={Proceedings of the IEEE/CVF conference on computer vision and pattern recognition},
  pages={9592--9600},
  year={2019}
}

@inproceedings{visa,
  title={Spot-the-difference self-supervised pre-training for anomaly detection and segmentation},
  author={Zou, Yang and Jeong, Jongheon and Pemula, Latha and Zhang, Dongqing and Dabeer, Onkar},
  booktitle={European Conference on Computer Vision},
  pages={392--408},
  year={2022},
  organization={Springer}
}

@inproceedings{realiad,
  title={Real-iad: A real-world multi-view dataset for benchmarking versatile industrial anomaly detection},
  author={Wang, Chengjie and Zhu, Wenbing and Gao, Bin-Bin and Gan, Zhenye and Zhang, Jiangning and Gu, Zhihao and Qian, Shuguang and Chen, Mingang and Ma, Lizhuang},
  booktitle={Proceedings of the IEEE/CVF Conference on Computer Vision and Pattern Recognition},
  pages={22883--22892},
  year={2024}
}

@inproceedings{padim,
  title={Padim: a patch distribution modeling framework for anomaly detection and localization},
  author={Defard, Thomas and Setkov, Aleksandr and Loesch, Angelique and Audigier, Romaric},
  booktitle={International conference on pattern recognition},
  pages={475--489},
  year={2021},
  organization={Springer}
}

@inproceedings{patchcore,
  title={Towards total recall in industrial anomaly detection},
  author={Roth, Karsten and Pemula, Latha and Zepeda, Joaquin and Sch{\"o}lkopf, Bernhard and Brox, Thomas and Gehler, Peter},
  booktitle={Proceedings of the IEEE/CVF conference on computer vision and pattern recognition},
  pages={14318--14328},
  year={2022}
}

@article{uniad,
  title={A unified model for multi-class anomaly detection},
  author={You, Zhiyuan and Cui, Lei and Shen, Yujun and Yang, Kai and Lu, Xin and Zheng, Yu and Le, Xinyi},
  journal={Advances in Neural Information Processing Systems},
  volume={35},
  pages={4571--4584},
  year={2022}
}

@inproceedings{omnial,
  title={Omnial: A unified cnn framework for unsupervised anomaly localization},
  author={Zhao, Ying},
  booktitle={Proceedings of the IEEE/CVF Conference on Computer Vision and Pattern Recognition},
  pages={3924--3933},
  year={2023}
}

@article{hvq-trans,
  title={Hierarchical vector quantized transformer for multi-class unsupervised anomaly detection},
  author={Lu, Ruiying and Wu, YuJie and Tian, Long and Wang, Dongsheng and Chen, Bo and Liu, Xiyang and Hu, Ruimin},
  journal={Advances in Neural Information Processing Systems},
  volume={36},
  pages={8487--8500},
  year={2023}
}

@inproceedings{rd4ad,
  title={Anomaly detection via reverse distillation from one-class embedding},
  author={Deng, Hanqiu and Li, Xingyu},
  booktitle={Proceedings of the IEEE/CVF conference on computer vision and pattern recognition},
  pages={9737--9746},
  year={2022}
}

@inproceedings{simplenet,
  title={Simplenet: A simple network for image anomaly detection and localization},
  author={Liu, Zhikang and Zhou, Yiming and Xu, Yuansheng and Wang, Zilei},
  booktitle={Proceedings of the IEEE/CVF conference on computer vision and pattern recognition},
  pages={20402--20411},
  year={2023}
}

@inproceedings{destseg,
  title={Destseg: Segmentation guided denoising student-teacher for anomaly detection},
  author={Zhang, Xuan and Li, Shiyu and Li, Xi and Huang, Ping and Shan, Jiulong and Chen, Ting},
  booktitle={Proceedings of the IEEE/CVF Conference on Computer Vision and Pattern Recognition},
  pages={3914--3923},
  year={2023}
}

@inproceedings{diad,
  title={A diffusion-based framework for multi-class anomaly detection},
  author={He, Haoyang and Zhang, Jiangning and Chen, Hongxu and Chen, Xuhai and Li, Zhishan and Chen, Xu and Wang, Yabiao and Wang, Chengjie and Xie, Lei},
  booktitle={Proceedings of the AAAI conference on artificial intelligence},
  volume={38},
  number={8},
  pages={8472--8480},
  year={2024}
}

@article{mambaad,
  title={Mambaad: Exploring state space models for multi-class unsupervised anomaly detection},
  author={He, Haoyang and Bai, Yuhu and Zhang, Jiangning and He, Qingdong and Chen, Hongxu and Gan, Zhenye and Wang, Chengjie and Li, Xiangtai and Tian, Guanzhong and Xie, Lei},
  journal={arXiv preprint arXiv:2404.06564},
  year={2024}
}

@article{inp,
  title={Exploring Intrinsic Normal Prototypes within a Single Image for Universal Anomaly Detection},
  author={Luo, Wei and Cao, Yunkang and Yao, Haiming and Zhang, Xiaotian and Lou, Jianan and Cheng, Yuqi and Shen, Weiming and Yu, Wenyong},
  journal={arXiv preprint arXiv:2503.02424},
  year={2025}
}

@article{dinomaly,
  title={Dinomaly: The Less Is More Philosophy in Multi-Class Unsupervised Anomaly Detection},
  author={Guo, Jia and Lu, Shuai and Zhang, Weihang and Chen, Fang and Liao, Hongen and Li, Huiqi},
  journal={arXiv preprint arXiv:2405.14325},
  year={2024}
}

@article{prototypelearning,
  title={Prototypical networks for few-shot learning},
  author={Snell, Jake and Swersky, Kevin and Zemel, Richard},
  journal={Advances in neural information processing systems},
  volume={30},
  year={2017}
}

@inproceedings{focal,
  title={Focal loss for dense object detection},
  author={Lin, Tsung-Yi and Goyal, Priya and Girshick, Ross and He, Kaiming and Doll{\'a}r, Piotr},
  booktitle={Proceedings of the IEEE international conference on computer vision},
  pages={2980--2988},
  year={2017}
}

@article{dinov2,
  title={Vision transformers need registers},
  author={Darcet, Timoth{\'e}e and Oquab, Maxime and Mairal, Julien and Bojanowski, Piotr},
  journal={arXiv preprint arXiv:2309.16588},
  year={2023}
}

@article{loco,
  title={Beyond dents and scratches: Logical constraints in unsupervised anomaly detection and localization},
  author={Bergmann, Paul and Batzner, Kilian and Fauser, Michael and Sattlegger, David and Steger, Carsten},
  journal={International Journal of Computer Vision},
  volume={130},
  number={4},
  pages={947--969},
  year={2022},
  publisher={Springer}
}

@article{mamba,
  title={Mamba: Linear-time sequence modeling with selective state spaces},
  author={Gu, Albert and Dao, Tri},
  journal={arXiv preprint arXiv:2312.00752},
  year={2023}
}

@inproceedings{memorizing,
  title={Memorizing normality to detect anomaly: Memory-augmented deep autoencoder for unsupervised anomaly detection},
  author={Gong, Dong and Liu, Lingqiao and Le, Vuong and Saha, Budhaditya and Mansour, Moussa Reda and Venkatesh, Svetha and Hengel, Anton van den},
  booktitle={Proceedings of the IEEE/CVF international conference on computer vision},
  pages={1705--1714},
  year={2019}
}

@inproceedings{pixel,
  title={Pixel-level anomaly detection via uncertainty-aware prototypical transformer},
  author={Huang, Chao and Liu, Chengliang and Zhang, Zheng and Wu, Zhihao and Wen, Jie and Jiang, Qiuping and Xu, Yong},
  booktitle={Proceedings of the 30th ACM international conference on multimedia},
  pages={521--530},
  year={2022}
}

@inproceedings{learning,
  title={Learning normal dynamics in videos with meta prototype network},
  author={Lv, Hui and Chen, Chen and Cui, Zhen and Xu, Chunyan and Li, Yong and Yang, Jian},
  booktitle={Proceedings of the IEEE/CVF conference on computer vision and pattern recognition},
  pages={15425--15434},
  year={2021}
}

@inproceedings{park2020learning,
  title={Learning memory-guided normality for anomaly detection},
  author={Park, Hyunjong and Noh, Jongyoun and Ham, Bumsub},
  booktitle={Proceedings of the IEEE/CVF conference on computer vision and pattern recognition},
  pages={14372--14381},
  year={2020}
}

@article{wideresnet,
  title={Wide residual networks},
  author={Zagoruyko, Sergey and Komodakis, Nikos},
  journal={arXiv preprint arXiv:1605.07146},
  year={2016}
}

@article{mdnd,  
 title={Modeling the Distribution of Normal Data in Pre-Trained Deep Features for Anomaly Detection}, 
 journal={Cornell University - arXiv,Cornell University - arXiv}, 
 author={Rippel, Oliver and Mertens, Patrick and Merhof, Dorit}, 
 year={2020}, 
 month={May}, 
 language={en-US} 
 }

@article{dfm,   title={Probabilistic Modeling of Deep Features for Out-of-Distribution and Adversarial Detection},  journal={Cornell University - arXiv,Cornell University - arXiv},  author={Ahuja, Nilesh and Ndiour, IbrahimaJ. and Kalyanpur, Trushant and Tickoo, Omesh},  year={2019},  month={Sep},  language={en-US}  }

@inproceedings{prototype2,
  title={Adaptive prototype learning and allocation for few-shot segmentation},
  author={Li, Gen and Jampani, Varun and Sevilla-Lara, Laura and Sun, Deqing and Kim, Jonghyun and Kim, Joongkyu},
  booktitle={Proceedings of the IEEE/CVF conference on computer vision and pattern recognition},
  pages={8334--8343},
  year={2021}
}

\end{document}